\documentclass{article}
\usepackage{graphicx} % Required for inserting images
\usepackage[numbers]{natbib}
\setcitestyle{square}
\usepackage{url}
\usepackage{fancyhdr}
\usepackage[draft]{changes}
\usepackage{geometry}
\usepackage{hyperref}
\usepackage[misc]{ifsym}
\usepackage{authblk}%Usage for this package: https://mirrors.ibiblio.org/CTAN/macros/latex/contrib/changes/changes.english.pdf

\title{\vspace{-2ex}\vspace{-2ex}Evaluation of Large Language Models for Summarization Tasks in the Medical Domain: A Narrative Review\vspace{-1ex}}
\author{Emma Croxford$^{1}$, Yanjun Gao PhD$^{2}$, Nicholas Pellegrino$^{3}$, Karen K. Wong MD MPH MIDS$^{3}$, Graham Wills PhD$^{4}$, Elliot First$^{3}$, Frank J. Liao PhD$^{1, 4}$, Cherodeep Goswami MBA$^{4}$, Brian Patterson MD MPH$^{1,4}$, Majid Afshar MD MSCR\href{mailto:mafshar@medicine.wisc.edu}{\Letter}} 
\date{}

\affil[1]{University of Wisconsin, Madison, USA}
\affil[2]{University of Colorado, Aurora, USA}
\affil[3]{Epic Systems, Verona, WI, USA}
\affil[4]{UW Health, Madison, WI, USA\vspace{-1ex}\vspace{-1ex}}
\begin{document}

\maketitle
\section{Abstract}
% Abstract is up to 70 words in length (~length of first paragraph)

\textit{Large Language Models have advanced clinical Natural Language Generation, creating opportunities to manage the volume of medical text. However, the high-stakes nature of medicine requires reliable evaluation, which remains a challenge. In this narrative review, we assess the current evaluation state for clinical summarization tasks and propose future directions to address the resource constraints of expert human evaluation.}

% Main Text is typically 3,000-4,000 words in length
\section{Introduction}

The rapid development of Large Language Models (LLMs) has led to significant advancements in the field of Natural Language Generation (NLG). In the medical domain, LLMs have shown promise in reducing documentation-based cognitive burden for healthcare providers, particularly in NLG tasks such as summarization and question answering. Summarizing clinical documentation has emerged as a critical NLG task as the volume of medical text in Electronic Health Records (EHRs) continues to expand ~\citep{Patterson_Hekman_Liao_Hamedani_Shah_Afshar_2024}.

Recent advancements, like the introduction of larger context windows in LLMs (e.g., Google’s Gemini 1.5 Pro with a 1 million-token capacity~\citep{GeminiTeam}), allow for the processing of extensive textual data, making it possible to summarize entire patient histories in a single input. However, a major challenge in applying LLMs to high-stakes environments like medicine is ensuring the reliable evaluation of their performance. Unlike traditional approaches, generative AI (GenAI) offers greater flexibility by generating natural language narratives that use language dynamically to fulfill tasks. Yet, this flexibility introduces added complexity in assessing the accuracy, reliability, and quality of the generated output where the desired response is not as static.

The evaluation of clinical summarization by LLMs must address the intricacies of complex medical texts and tackle LLM-specific challenges such as relevancy, hallucinations, omissions, and ensuring factual accuracy ~\citep{Zhao_Zhou_Li_Tang_Wang_Hou_Min_Zhang_Zhang_Dong_etal._2023}. Healthcare data can further complicate the LLM-specific challenges because they can contain conflicting or incorrect information. Current metrics, like n-gram overlap and semantic scores, used in summarization tasks are insufficient for the nuanced needs of the medical domain ~\citep{Moramarco_PapadopoulosKorfiatis_Perera_Juric_Flann_Reiter_Belz_Savkov_2022}. While these metrics may perform adequately for simple extractive summarization, they fall short when applied to abstractive summarization ~\citep{Croxford_Gao_Patterson_To_Tesch_Dligach_Mayampurath_Churpek_Afshar_2024}, where complex reasoning and in-depth medical knowledge are required. They are also unable to differentiate in the needs of various users and provide evaluations that account for the relevancy of generations.

In the era of GenAI, automation bias further complicates the potential risks posed by LLMs, particularly in clinical settings where the consequences of inaccuracies can be severe. Therefore, efficient and automated evaluation methods are essential. In this review, we examine the current state of LLM evaluation in summarization tasks, highlighting both its applications and limitations in the medical domain. We also propose a future direction to overcome the labor-intensive process of expert human evaluation, which is time-consuming, costly, and requires specialized training.

\section{Human Evaluations in Electronic Health Record Documentation}

The current human evaluation frameworks for human-authored clinical notes are largely based on pre-GenAI rubrics that assess clinical documentation quality. These frameworks vary depending on the type of evaluators, content, and the analysis required to generate evaluative scores. Such flexibility allows for tailored evaluation methods, capturing task-specific aspects that ensure quality generation. Expert evaluators, with their field-specific knowledge, play a crucial role in maintaining high standards of assessment. 

Some commonly used pre-GenAI rubrics include the SaferDx ~\citep{Singh_Khanna_Spitzmueller_Meyer_2019}, Physician Documentation Quality Instrument (PDQI-9) ~\citep{Stetson_Bakken_Wrenn_Siegler_2012}, and Revised-IDEA ~\citep{Schaye_Miller_Kudlowitz_Chun_Burk-Rafel_Cocks_Guzman_Aphinyanaphongs_Marin_2022} rubrics. The SaferDx rubric focuses on identifying diagnostic errors and analyzing missed opportunities in EHR documentation through a 12-question retrospective survey aimed at improving diagnostic decision-making and patient safety. The PDQI-9 evaluates physician note quality across nine criteria questions, ensuring continuous improvement in clinical documentation and patient care. The Revised-IDEA tool offers feedback on clinical reasoning documentation through a 4-item assessment. All three of these rubrics place emphasis on the omission of relevant diagnoses throughout the differential diagnosis process and the relevant objective data, processes, and conclusions associated with those diagnoses. They also require clinical documentation to be free of incorrect, inappropriate, or incomplete information emphasizing the importance of the quality of evidence and reasoning that is present in clinical documentation. Each rubric includes additional questions based on the origin and usage of specific clinical documentation — like the PDQI-9's assessment of organization to ensure a reader is able to understand the clinical course of a patient. Each of the three also uses different assessment styles based on the granularity of the questions and intention behind the assessment. For instance, the Revised-IDEA tool uses a count style assessment for 3 of the 4-items to guarantee the inclusion of a minimum number of objective data points and inclusion of required features for a high-quality diagnostic reasoning documentation. In recent publications, the SaferDx tool has been used as a retrospective analysis of the use of GenAI in clinical practice ~\citep{Kawamura_Harada_Sugimoto_Nagase_Katsukura_Shimizu_2022}, whereas the PDQI-9 and Revised-IDEA tools have been utilized to compare the quality of clinical documentation that is written by clinicians versus GenAI methods ~\citep{Tierney_Gayre_Hoberman_Mattern_Ballesca_Kipnis_Liu_Lee_2024, Eshel_Bellolio_Boggust_Shapiro_Mullan_Heaton_Madsen_Homme, Cabral_Restrepo_Kanjee_Wilson_Crowe_Abdulnour_Rodman_2024}. While each of these rubrics was not originally designed to evaluate LLM-generated content, they offer valuable insights into the essential criteria for evaluating text generated in the medical domain.

Human evaluations remain the gold standard for LLM outputs ~\citep{Sai_Mohankumar_Khapra_2023}. However, because these rubrics were initially developed for evaluating clinician-generated notes, they may need to be adapted for the specific purpose of evaluating LLM-generated output. Several new and modified evaluation rubrics have emerged to address the unique challenges posed by LLM-generated content, including evaluating the consistency and factual accuracy (i.e., hallucinations) of the generated text. Common themes in these adapted rubrics include safety ~\citep{Singhal_Azizi_Tu_Mahdavi_Wei_Chung_Scales}, modality ~\citep{Otmakhova_Verspoor_Baldwin_Lau_2022, Adams_Zucker_Elhadad_2023}, and correctness ~\citep{Guo_Qiu_Wang_Cohen_2022, Wallace_Saha_Soboczenski_Marshall_2020}.

\subsection{Criteria for Human Evaluations}

In general, the criteria that are used to make up evaluation rubrics for LLM output fall into seven broad criteria: (1) \textbf{Hallucination} ~\citep{Moramarco_PapadopoulosKorfiatis_Perera_Juric_Flann_Reiter_Belz_Savkov_2022, Guo_Qiu_Wang_Cohen_2022, Wallace_Saha_Soboczenski_Marshall_2020, Abacha_Yim_Michalopoulos_Lin_2023, Yadav_Gupta_Abacha_Demner-Fushman_2021, Moor_Huang_Wu_Yasunaga_Zakka_Dalmia_Reis_Rajpurkar_Leskovec_2023, Dalla_Serra_Clackett_MacKinnon_Wang_Deligianni_Dalton_O’Neil_2022}, (2) \textbf{Omission} ~\citep{Singhal_Azizi_Tu_Mahdavi_Wei_Chung_Scales, Abacha_Yim_Michalopoulos_Lin_2023}, (3) \textbf{Revision} ~\citep{Cai_Liu_Bajracharya_Sills_Kapoor_Liu_Berlowitz_Levy_Pradhan_Yu_2022}, (4) \textbf{Faithfulness/Confidence} ~\citep{Otmakhova_Verspoor_Baldwin_Lau_2022, Adams_Zucker_Elhadad_2023, Cai_Liu_Bajracharya_Sills_Kapoor_Liu_Berlowitz_Levy_Pradhan_Yu_2022}, (5) \textbf{Bias/Harm} ~\citep{Singhal_Azizi_Tu_Mahdavi_Wei_Chung_Scales, Adams_Zucker_Elhadad_2023, Dalla_Serra_Clackett_MacKinnon_Wang_Deligianni_Dalton_O’Neil_2022}, (6) \textbf{Groundedness} ~\citep{Singhal_Azizi_Tu_Mahdavi_Wei_Chung_Scales, Otmakhova_Verspoor_Baldwin_Lau_2022}, and (7) \textbf{Fluency} ~\citep{Otmakhova_Verspoor_Baldwin_Lau_2022, Guo_Qiu_Wang_Cohen_2022, Yadav_Gupta_Abacha_Demner-Fushman_2021, Cai_Liu_Bajracharya_Sills_Kapoor_Liu_Berlowitz_Levy_Pradhan_Yu_2022}. \textbf{Hallucination} encompasses any evaluative questions that intend to capture when information in a generated text does not follow from the source material. Unsupported claims, nonsensical statements, improbable scenarios, and incorrect or contradictory facts would be flagged by questions in this criteria. \textbf{Omission}-based questions are used to identify missing information in a generated text. Medical facts, important information, and critical diagnostic decisions can all be considered omitted when not included in generated text, if those items would have been included by a medical professional. When an evaluator is asked to make revisions or estimate the number of revisions needed for a generated text, the evaluative question would fall under \textbf{Revision}. Generated texts are revised until they meet the standards set forth by a researcher, hospital system, or larger government body. \textbf{Faithfulness/Confidence} is generally characterized by questions that capture whether a generated text has preserved the content of the source text and presented conclusions that reflect the confidence and specificity present in the source text. Questions about \textbf{Bias/Harm} evaluate whether generated text is introducing potential harm to a patient or reflecting bias in the response. Information that is inaccurate, inapplicable, or poorly applied would be captured by questions that fall under this criteria. \textbf{Groundedness} refers to evaluative questions that grade the quality of the source-based evidence for a generated text. Any evidence that contains poor reading comprehension, recall of knowledge, reasoning steps, or is antithetical to scientific consensus would result in a poor groundedness score. In addition to the content of a generated text, the \textbf{Fluency} of a generated text is also included in evaluations. Coherency, readability, grammatical correctness, and lexical correctness fall under this criteria. In many cases, Fluency is assumed to be adequate in favor of focusing on content-based evaluative criteria.

\subsection{Analysis of Human Evaluations}

The method of analysis for evaluation rubrics can also vary based upon the setting and task. Evaluative scores can be calculated using binary/Likert categorizations ~\citep{Singhal_Azizi_Tu_Mahdavi_Wei_Chung_Scales, Otmakhova_Verspoor_Baldwin_Lau_2022}, counts/proportions of pre-specified instances ~\citep{Dalla_Serra_Clackett_MacKinnon_Wang_Deligianni_Dalton_O’Neil_2022}, edit distance ~\citep{Cai_Liu_Bajracharya_Sills_Kapoor_Liu_Berlowitz_Levy_Pradhan_Yu_2022}, or penalty/reward schemes similar to those used for medical exams ~\citep{Umapathi_Pal_Sankarasubbu_2023}. \textbf{Binary} categorizations answer evaluative questions using True/False or Yes/No response schema. This set-up allows complex evaluations to be broken down into simpler and potentially more objective decisions. A binary categorization places more penalization on smaller errors by pushing responses to be either acceptable or unacceptable. \textbf{Likert}-scaled categorizations allow for a higher level of specificity in the score by providing an ordinal scale. These scales can consist of as many levels as necessary, and in many cases there are between 3 and 9 levels including a neutral option for unclear responses. Scales with a higher number of levels introduce more problems with meeting assumptions of a normal distribution into an analysis, along with complexity and disagreement amongst reviewers. \textbf{Count/proportion}-based evaluations require an evaluator to identify pre-specified instances of correct or incorrect key phrases related to a particular evaluative criteria. A precision, recall, f-score, or rate can then be computed from an evaluator's annotations to establish a numerical score for a generated text. \textbf{Edit distance} evaluations also require an evaluator to make annotations on the generated text that is being evaluated. In these cases, an evaluator makes edits to the generated text until it is satisfactory or no longer contains critical errors. These edits can be corrections on factual errors, inclusion of omissions, or removal of irrelevant items. The evaluative score is the distance from the original generated text and the edited version based upon the number of characters, words, etc. that required editing. The Levenshtein distance ~\citep{1966} is an example of an algorithm used to calculate the distance between the generated text and its edited version. This distance is calculated as the minimum number of substitutions, insertions, and deletions of individual characters required to change the original to the edited version. Finally, one of the more complex ways to compute evaluative scores is to use a \textbf{Penalty/Reward} schema. These schema award points for positive outcomes to evaluative questions and penalize negative outcomes. This schema is similar to those seen on national exams which account for positive and negative scores, using the importance and difficulty associated with different questions. For example, the schema used to evaluate LLMs on the Med-HALT dataset is an average of the correct and incorrect answers which are assigned +1 and -0.25 points respectively ~\citep{Umapathi_Pal_Sankarasubbu_2023}. This evaluation schema provides a high level of specificity for assigning weights representative of the trade-off between false positives and false negatives.  

\subsection{Drawbacks of Human Evaluations}

While human evaluations provide nuanced assessments, they are resource-intensive and heavily reliant on the recruitment of evaluators with clinical domain knowledge. The experience and background of an evaluator can significantly influence how they interpret and evaluate generated text. Additionally, the level of guidance and specificity in evaluative instructions determines how much of the assessment is shaped by the evaluators' personal interpretations and beliefs about the task. Although increasing the number of evaluators could mitigate some of these biases, resources—both time and financial—often limit the scale of human evaluations. These evaluations also require substantial manual effort, and without clear guidelines and training, inter-rater agreement may suffer. Ensuring that human evaluators align with the evaluation rubric’s intent requires training, much like annotation guidelines for NLP shared tasks ~\citep{Gao_Dligach_Miller_Tesch_Laffin_Churpek_Afshar_2022, Goldsack_Scarton_Shardlow_Lin_2024, Gupta_Demner-Fushman_2022}. In the clinical domain, medical professionals are typically used as expert evaluators, but their time constraints limit their availability for large-scale evaluations. The difficulty of recruiting more medical professionals, compounded by the time needed for thorough assessments, makes frequent, rapid evaluations impractical.

Another concern is the validity of the evaluation rubric itself. A robust human evaluation framework must possess strong psychometric properties, including construct validity, criterion validity, content validity, and inter-rater reliability, to ensure reproducibility and generalizability. Unfortunately, many frameworks used in clinical evaluations do not provide sufficient details about their creation, making it difficult to assess their validity ~\citep{Otmakhova_Verspoor_Baldwin_Lau_2022, Umapathi_Pal_Sankarasubbu_2023}. Often, human evaluation frameworks are developed for specific projects with only one evaluator, and while metrics like inter-rater reliability are crucial to establish validity, they are not always reported ~\citep{Wallace_Saha_Soboczenski_Marshall_2020, Cai_Liu_Bajracharya_Sills_Kapoor_Liu_Berlowitz_Levy_Pradhan_Yu_2022}. Moreover, clinically relevant evaluation rubrics have not been specifically designed to assess LLM-generated summaries. Most existing evaluation rubrics focus on assessing human-authored note quality, and they do not encompass all the elements required to evaluate the unique aspects of LLM-generated outputs ~\citep{Singh_Khanna_Spitzmueller_Meyer_2019, Stetson_Bakken_Wrenn_Siegler_2012, Schaye_Miller_Kudlowitz_Chun_Burk-Rafel_Cocks_Guzman_Aphinyanaphongs_Marin_2022}.

\section{Pre-LLM Automated Evaluations}

Automated metrics offer a practical solution to the resource constraints of human evaluations, particularly in fields like Natural Language Processing (NLP), where tasks such as question answering, translation, and summarization have long relied on these methods. Automated evaluations employ algorithms, models, or heuristic techniques to assess the quality of generated text without the need for continuous human intervention, making them far more efficient in terms of time and labor. These metrics, however, depend heavily on the availability of high-quality reference texts, often referred to as "gold standards." The generated text is compared against these gold standard reference texts to evaluate its accuracy and how well it meets the task's requirements. Despite their efficiency, automated metrics may struggle to capture the nuance and contextual understanding required in more complex domains, such as clinical diagnosis, where subtle differences in phrasing or reasoning can have significant implications. Therefore, while automated evaluations are valuable for their scalability, their effectiveness is closely tied to the quality and relevance of the reference texts used in the evaluation.

\subsection{Categories of Automated Evaluation}

\begin{figure} [ht]
    \centering
    \includegraphics[scale = 0.2, angle = 90]{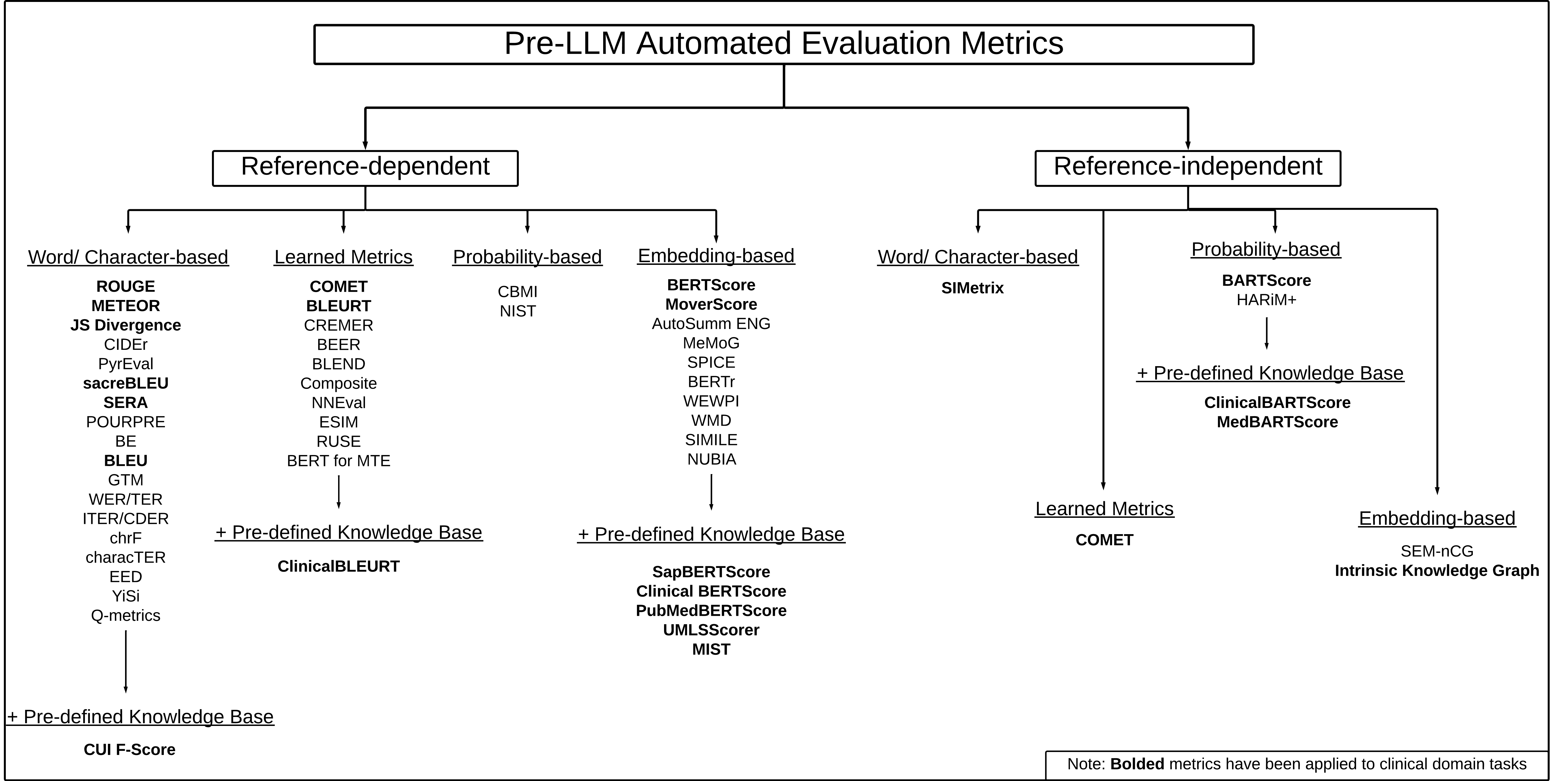}
    \caption{\textbf{Pre-LLM Automated Evaluation Metric Taxonomy} A structured organization of pre-LLM automated evaluation metrics categorized by their bases and the need for ground truth references. Those metrics that were built for or have been applied in the clinical domain are in bold.\protect\footnotemark[1] }
    \label{fig:taxonomy}
\end{figure}

Automated evaluations in the clinical domain can be categorized into five primary types (Figure \ref{fig:taxonomy}), each tailored to specific evaluation goals and dependent on the availability of reference and source material for the generated text: (1) \textbf{Word/Character-based}, (2) \textbf{Embedding-based}, (3) \textbf{Learned metrics}, (4) \textbf{Probability-based}, (5) and \textbf{Pre-Defined Knowledge Base}. 

\textbf{Word/Character}-based evaluations rely on comparisons between a reference text and the generated text to compute an evaluative score. These evaluations can be based on character, word, or sub-sequence overlaps depending on the need of the evaluation and the nuance that may be present in the text. Recall Oriented Understudy for Gisting Evaluation (ROUGE) ~\citep{Lin} is a prime example of a word/character-based metric. The many variants of ROUGE — N-gram Co-Occurrence (N), Longest Common Sub-sequence (L), Weighted Longest Common Sub-sequence (W), Skip-Bigram Co-Occurrence (S) — represent the level of comparison between the reference and generated texts. ROUGE-L is the current gold standard for automated evaluation, especially in summarization, and relies on the longest common subsequence between the reference and generated texts. The evaluative score is computed as the fraction of words in the text that are in the longest common subsequence. Edit distance metrics ~\citep{1966} would also fall under this category as they are based on the number of words or characters that would need to be changed to match the reference and generated texts. Edits can be classified as insertions, deletions, substitutions, or transpositions of the words/characters in the generated text.

\textbf{Embedding-based} evaluations create contextualized or static embeddings for the reference and generated texts for comparison rather than relying on exact matches between words or characters. These embedding-based metrics are able to capture semantic similarities between two texts since the embedding for a word or phrase would be based on the text that surrounds it as well as itself. The BERTScore ~\citep{Zhang_Kishore_Wu_Weinberger_Artzi_2020a} is a commonly used metric that falls under this category. For this metric, a Bidirectional Encoder Representations from Transformers (BERT) model ~\citep{Devlin_Chang_Lee_Toutanova_2019} is used to generate the contextualized embeddings before computing a greedy cosine similarity score based on those embeddings. 

\newgeometry{top=0.2in}

\restoregeometry

 \footnotetext[1]{ 
    The taxonomy includes 
    Recall-Oriented Understudy for Gisting Evaluation (ROUGE)~\citep{Lin}, 
    Metric for Evaluation of Translation with Explicit Ordering (METEOR)~\citep{METEOR}, 
    Jensen–Shannon (JS) Divergence~\citep{Louis_Nenkova_2013}, 
    Consensus-based Image Description Evaluation (CIDEr)~\citep{CIDEr}, 
    PyrEval~\citep{Gao_Sun_Passonneau}, 
    Standardized Bilingual Evaluation Understudy (sacreBLEU)~\citep{BLEU}, 
    Summarization Evaluation by Relevance Analysis (SERA)~\citep{Cohan_Goharian}, 
    POURPRE~\citep{Lin_Demner-Fushman_2005}, 
    Basic Elements (BE)~\citep{Hovy_Lin_Zhou_Fukumoto_2006}, 
    Bilingual Evaluation Understudy (BLEU)~\citep{BLEU}, 
    General Text Matcher (GTM)~\citep{Turian_Shen_Melamed_2003}, 
    Word Error Rate (WER)~\citep{Su_Wu_Chang_1992}/ Translation Edit Rate (TER)~\citep{Snover_Dorr_Schwartz_Micciulla_Makhoul_2006}, 
    Improving Translation Edit Rate (ITER)~\citep{ITER}/CDER (Cover-Disjoint Error Rate)~\citep{CDER}, 
    chrF (character n-gram F-score)~\citep{chrF}, 
    characTER (Character Level Translation Edit Rate)~\citep{CharacTer}, 
    Extended Edit Distance (EED)~\citep{EED}, 
    YiSi ~\citep{YiSi}, %(意思)
    Q-metrics~\citep{Nema_Khapra_2018}, 
    Concept Unique Identifier (CUI) F-Score~\citep{Gao_Dligach_Miller_Xu_Churpek_Afshar_2022}, 
    Crosslingual Optimized Metric for Evaluation of Translation (COMET)~\citep{Rei_Stewart_Farinha_Lavie_2020}, 
    Bilingual Evaluation Understudy with Representations from Transformers (BLEURT)~\citep{BLEURT}, 
    Combined Regression Model for Evaluating Responsiveness (CREMER)~\citep{Lin_Liu_Ng_Kan_2012}, 
    Better Evaluation as Ranking (BEER)~\citep{Stanojević_Sima’an_2014}, 
    BLEND~\citep{Ma_Graham_Wang_Liu_2017}, 
    Composite~\citep{Sharif_White_Bennamoun_Ali_Shah_2018}, 
    Neural Network Based Evaluation Metric (NNEval)~\citep{Sharif_White_Bennamoun_Ali_Shah_2018}, 
    Enhanced Sequential Inference Model (ESIM)~\citep{Chen_Zhu_Ling_Wei_Jiang_Inkpen_2017}, 
    Regressor Using Sentence Embeddings (RUSE)~\citep{RUSE}, 
    Bidirectional Encoder Representations from Transformers for Machine Translation Evaluation (BERT for MTE)~\citep{Shimanaka_Kajiwara_Komachi_2019}, 
    ClinicalBLEURT~\citep{Abacha_Yim_Michalopoulos_Lin_2023}, 
    Conditional Bilingual Mutual Information (CBMI)~\citep{Zhang_Liu_Meng_Chen_Xu_Liu_Zhou_2022}, 
    NIST~\citep{Doddington_2002}, 
    BERTScore~\citep{Zhang_Kishore_Wu_Weinberger_Artzi_2020a}, 
    MoverScore~\citep{MoverScore}, 
    AUTOmatic SUMMary Evaluation based on N-gram Graphs (AutoSumm ENG)~\citep{Giannakopoulos_Karkaletsis}, 
    Merge Model Graph (MeMoG)~\citep{Giannakopoulos_Karkaletsis}, 
    Semantic Propositional Image Caption Evaluation (SPICE)~\citep{SPICE}, 
    BERTr~\citep{Mathur_Baldwin_Cohn_2019}, 
    Word Embedding-based automatic MT evaluation using Word Position Information (WEWPI)~\citep{WE_WPI}, 
    Word Mover-Distance (WMD)~\citep{WMD}, 
    SIMILE~\citep{sacreBLEU},   
    NeUral Based Interchangeability Assessor (NUBIA)~\citep{NUBIA}, 
    SapBERTScore~\citep{liu-etal-2021-self}, 
    ClinicalBERTScore~\citep{clinicalbert}, 
    PubMedBERTScore~\citep{pubmedbert}, 
    UMLSScorer~\citep{UMLSScorer-JB}, 
    MIST~\citep{Abacha_Yim_Michalopoulos_Lin_2023}, 
    Summary-Input Similarity Metrics (SIMetrix)~\citep{Louis_Nenkova_2013}, 
    BARTScore~\citep{Yuan_Neubig_Liu_2021}, 
    Hallucination Risk Measurement+ (HARiM+)~\citep{Son_Park_Hwang_Lee_Noh_Lee}, 
    ClinicalBARTScore~\citep{Yuan_Neubig_Liu_2021}, 
    MedBARTScore~\citep{Abacha_Yim_Michalopoulos_Lin_2023}, 
    Semantic Normalized Cumulative Gain (SEM-nCG)~\citep{Akter_Bansal_Karmaker_2022}, 
    Intrinsic Knowledge Graph~\citep{Aracena_Villena_Rojas_Dunstan}
    }

\textbf{Learned metric}-based evaluations rely on training a model to compute the evaluations. These metrics can be trained on example evaluation scores or directly on the reference and generated text pairs. Regression and neural network models are the foundation of these metrics providing varying degrees of complexity for the learnable parameters. The Crosslingual Optimized Metric for Evaluation of Translation (COMET) ~\citep{Rei_Stewart_Farinha_Lavie_2020} is a metric that would fall under this category as it is a neural model trained for evaluation. It was originally created for evaluation of machine translations, but has since been applied to other generative tasks. COMET uses a neural network with the generated text as input to produce an evaluative score. This metric can be applied to datasets that are reference-less as well as those with reference texts.

\textbf{Probability} evaluations rely on calculating the likelihood of a generated text based on domain knowledge, reference texts, or source material. These metrics equate high-quality generations with those that have a high probability of being coherent or relevant to the reference or source text. They also penalize the inclusion of off-topic or unrelated information. An example is BARTScore ~\citep{Yuan_Neubig_Liu_2021}, which calculates the sum of log probabilities for the generated output based on the reference text. In this case, the log probabilities are computed using the Bidirectional and Auto-Regressive Transformer (BART) model, which assesses how well the generated text aligns with the expected content ~\citep{Lewis_Liu_Goyal_Ghazvininejad_Mohamed_Levy_Stoyanov_Zettlemoyer_2020}. 

\textbf{Pre-Defined Knowledge Base} metrics rely on established databases of domain-specific knowledge to inform the evaluation of generated text. These metrics are particularly valuable in specialized fields like healthcare, where general language models may lack the necessary depth of knowledge. By incorporating domain-specific knowledge bases, such as the National Library of Medicine's Unified Medical Language System (UMLS) ~\citep{UMLS_Metathesaurus_Browser}, these metrics provide more accurate and contextually relevant evaluations. Pre-defined knowledge bases can enhance other evaluation methods, such as contextual embedding, machine learning, or probability-based metrics, by grounding them in the specialized terminology and relationships unique to the domain. This combination ensures that evaluations account for both linguistic accuracy and the specialized knowledge required in fields like clinical medicine. BERTScore has a variant that was trained on the UMLS called the SapBERTScore ~\citep{Liu_Shareghi_Meng_Basaldella_Collier_2021}. The score functions similarly to the general domain BERTScore but leverages a BERT model fine-tuned using UMLS data to generate more domain-specific embeddings. Other metrics based on the UMLS include the CUI F-Score ~\citep{Gao_Dligach_Miller_Xu_Churpek_Afshar_2022} and UMLS Scorer ~\citep{UMLSScorer-JB}. The UMLS Scorer utilizes UMLS-based knowledge graph embeddings to assess the semantic quality of the text ~\citep{Abacha_Yim_Michalopoulos_Lin_2023}, providing a more structured approach to evaluating clinical content. Meanwhile, the CUI F-Score represents text using Concept Unique Identifiers (CUIs) from the UMLS, calculating F-scores that reflect how well the generated text aligns with key medical concepts. This enables a more granular evaluation of the relevance and accuracy of medical terminology within the generated content.

\subsection{Drawbacks of Automated Metrics}

Prior to the advent of LLMs, automated metrics would generate a single score meant to represent the quality of a generated text, regardless of its length or complexity. This single-score approach can make it difficult to pinpoint specific issues in the text, and in the case of LLMs, it is nearly impossible to understand the precise factors contributing to a particular score ~\citep{Sai_Mohankumar_Khapra_2023}. While automated metrics offer the benefit of speed, this comes at the cost of relying on surface-level heuristics, such as lexicographic and structural measures, that fail to capture more abstract summarization challenges in medical text such as needing to apply clinical reasoning and knowledge to appropriately prioritize and synthesize medical information.

\section{FUTURE DIRECTIONS: LLMs as Evaluators to Complement Human Expert Evaluators: Prompt Engineering LLMs as \break Judges}

\begin{figure} [ht]
    \centering
    \includegraphics[width = \textwidth]{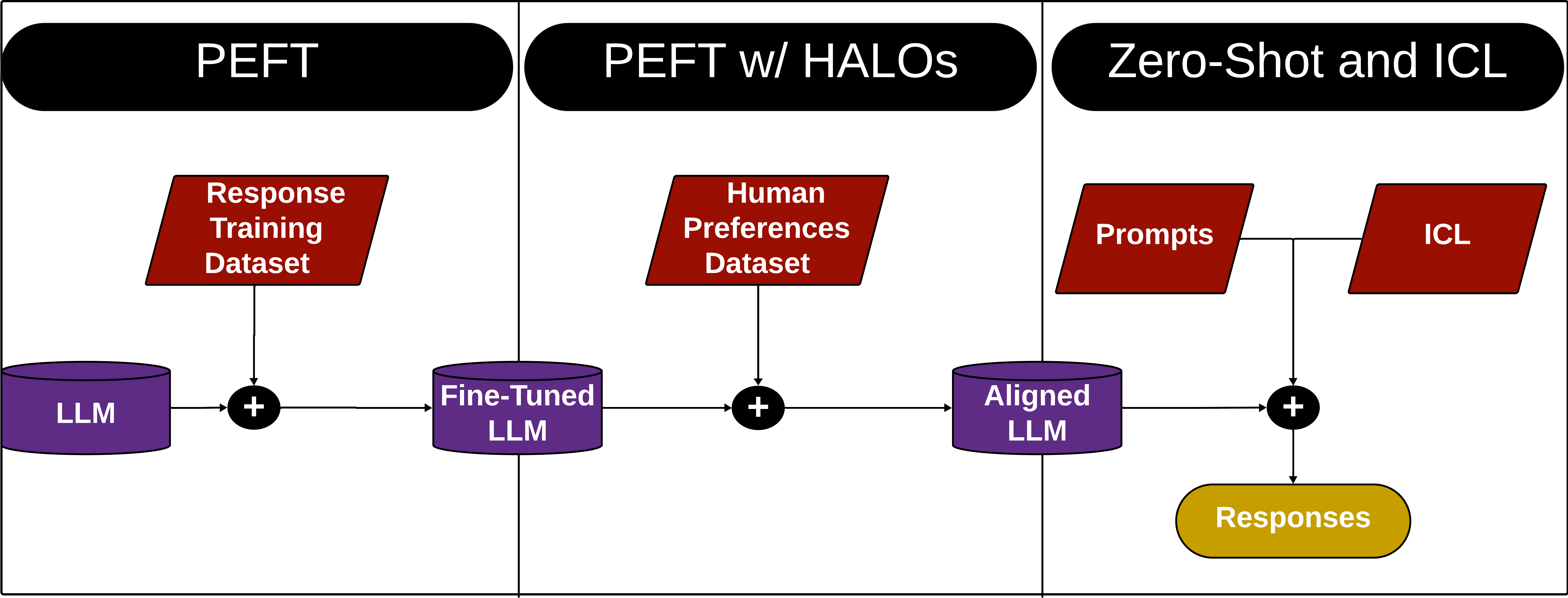}
    \caption{\textbf{Stages of Prompt Engineering LLMs as Judges} The three different aspects of prompt engineering expanded upon in section 5. The three sections - Zero-Shot and In-Context Learning (ICL), Parameter Efficient Fine Tuning (PEFT), and PEFT with Human Aware Loss Function (HALO) - fit together into a larger schema for training and prompting an LLM to serve as an evaluator to complement human expert evaluators. }
    \label{fig:overview}
\end{figure}

LLMs are versatile tools capable of performing a wide range of tasks, including evaluating the outputs of other LLMs. This concept, where an LLM acts as a model of a human expert evaluator, has gained traction with the advent of instruction tuning and reinforcement learning with human feedback (RLHF) ~\citep{Christiano_Leike_Brown_Martic_LeggAmodei2017}. These advancements have significantly improved the ability of LLMs to align their outputs with human preferences, as seen in the transition from GPT-3 to GPT-4, which marked a paradigm shift in LLM accuracy and performance ~\citep{OpenAI_Achiam_Adler_Agarwal_Ahmad_Akkaya_Aleman_Almeida_Altenschmidt_Altman_etal._2024}.

An effective LLM evaluator would be able to respond to evaluative questions with precision and accuracy comparable to that of human experts, following frameworks like those used in human evaluation rubrics. LLM-based evaluations could provide many of the same advantages as traditional automated metrics, such as speed and consistency, while potentially overcoming the reliance on high-quality reference texts. Moreover, LLMs could evaluate complex tasks by directly engaging with the content, bypassing the need for simplistic heuristics and offering more information into factual accuracy, hallucinations, and omissions.

Although the use of LLMs as evaluators is still emerging in research, early studies have demonstrated their utility as an alternative to human evaluations, offering a scalable solution to the limitations of manual assessment ~\citep{Zheng_Chiang_Sheng_Zhuang_Wu_Zhuang_Lin_Li_Li_Xing_etal._2023}. As the methodology continues to develop, LLM-based evaluations hold promise for addressing the shortcomings of both traditional automated metrics and human evaluations, particularly in complex, context-rich domains such as clinical text generation.

\subsection{Zero-Shot and In-Context Learning} 

\begin{figure} [ht]
    \centering
    \includegraphics[width = \textwidth]{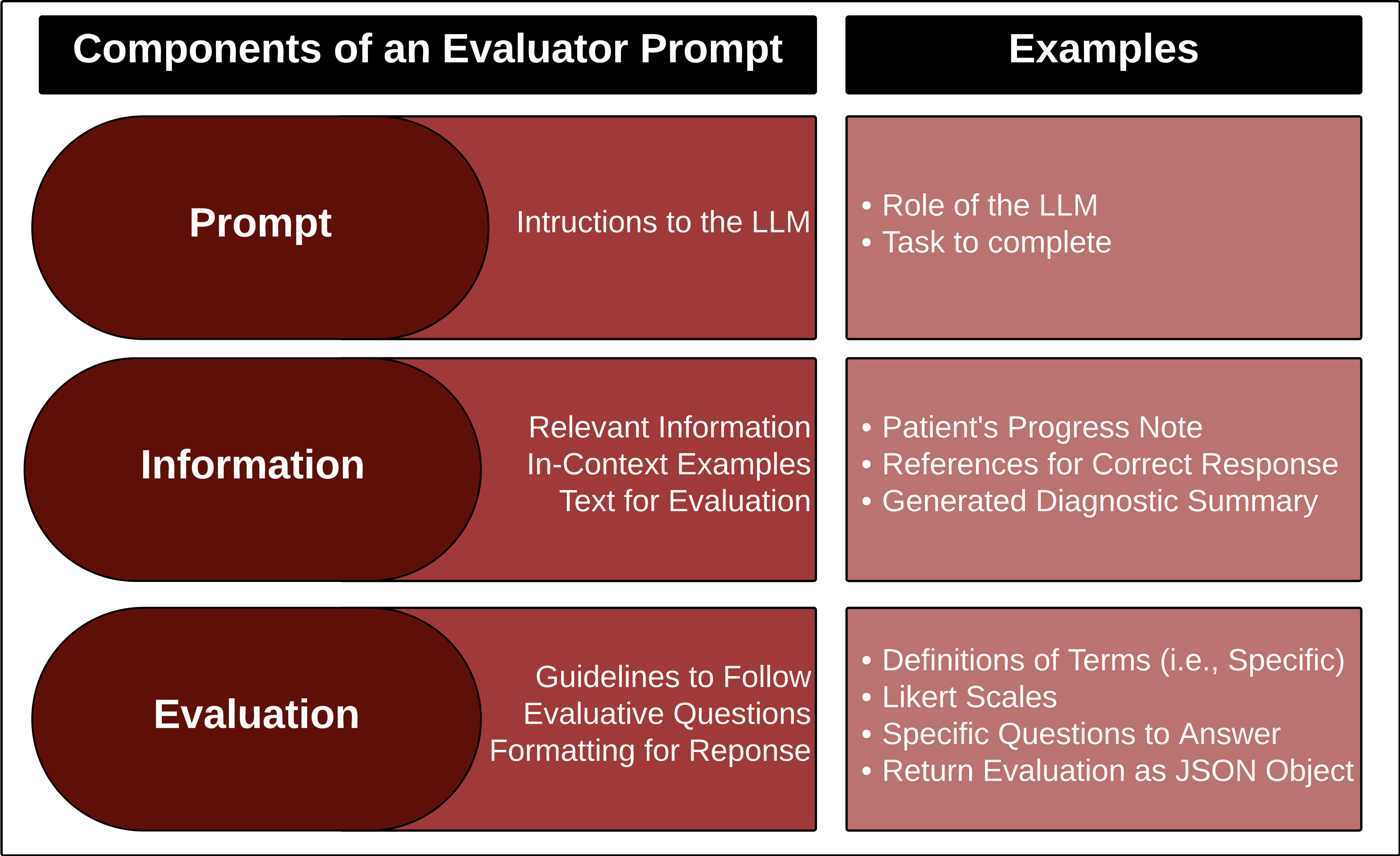}
    \caption{\textbf{Anatomy of an Evaluator Prompt} An evaluator prompt consists of three sections: Prompt, Information, and Evaluation. All three components are essential for an LLM serving as an evaluator. The Evaluator Prompt needs to instruct the LLM on the task (Prompt), provide the LLM will all the necessary information to make an evaluation (Information), and all the information that defines the guidelines and formatting of the evaluation (Evaluation).}
    \label{fig:prompts}
\end{figure}

One method for designing LLMs to perform evaluations is through the use of manually curated prompts (Figure ~\ref{fig:prompts}). A prompt consists of the task description and instructions provided to an LLM to guide its responses. Two primary prompting strategies are employed in this context: Zero-Shot and Few-Shot ~\citep{Zhao_Zhou_Li_Tang_Wang_Hou_Min_Zhang_Zhang_Dong_etal._2023}. In Zero-Shot prompting, the LLM is given only the task description without any examples before being asked to perform evaluations.  Few-Shot prompting provides the task description alongside a few examples to help guide the LLM in generating output. The number of examples varies based on the LLM's architecture, input window limitations, and the point at which the model performs optimally. Typically, between one and five few-shot examples are used. Prompt engineering, through both Zero-Shot and Few-Shot ("in-context learning") approaches (collectively referred to as "hard prompting"), enables an LLM to perform tasks that it was not explicitly trained to do. However, performance can vary significantly depending on the model’s pre-training and its relevance to the new task.

Beyond these manual approaches, a more adaptive strategy involves "soft prompting," also known as machine-learned prompts, which includes techniques like prompt tuning and p-tuning ~\citep{Lester_Al-Rfou_Constant_2021}. Soft prompts are learnable parameters added as virtual tokens to a model's input to signal task-specific instructions. Unlike hard prompts, soft prompts are trained and incorporated into the model’s input layer, enabling the model to handle a broader range of specialized tasks. Soft prompting has been shown to outperform Few-Shot prompting, especially in large-scale models, as it fine-tunes the model’s behavior without altering the core weights. When prompting alone does not achieve the desired performance, fine-tuning the entire LLM may be necessary for optimal task execution.

\subsection{Parameter Efficient Fine-Tuning}

Even though an LLM may be pre-trained on a vast corpus, it can struggle with tasks requiring domain-specific knowledge or handling nuanced inputs. To address these challenges, Supervised fine-tuning (SFT) methods with Parameter Efficient Fine-Tuning (PEFT) using quantization and low rank adaptors can be employed, where the model is trained on a specialized dataset of prompt/response pairs tailored to the task at hand. Fine-tuning every weight in a LLM can require a large amount of time and computational resources. In these instances, quantization and low rank adaptors are added to the fine-tuning process for PEFT. Quantization reduces the time and memory costs of training by using lower precision data types, generally 4-bit and 8-bit, for the LLMs weights ~\citep{Dettmers_Pagnoni_Holtzman_Zettlemoyer_2023}. Low rank adaptors (LoRA) freeze the weights of a LLM and decompose them into a smaller number of trainable parameters ultimately also reducing the costs of SFT ~\citep{Hu_Shen_Wallis_Allen-Zhu_Li_Wang_Wang_Chen_2021}. PEFT helps refine an LLM by embedding task-specific knowledge, ensuring the model can respond accurately in specialized contexts. The creation of these datasets is critical—performance improvements are directly tied to the quality and relevance of the prompt/response pairs used for fine-tuning. The goal is to adjust the LLM to perform better in specific use cases, such as medical diagnosis or legal reasoning, by narrowing its focus to task-specific behaviors through PEFT.  

\subsection{Parameter Efficient Fine-Tuning with Human-Aware Loss Function}

In certain applications, the focus of fine-tuning is to align the LLM with human values and preferences, especially when the model risks generating biased, incorrect, or harmful content. This alignment, known as Human Alignment training, is driven by high-quality human feedback integrated into the training process. A widely recognized approach in this domain is Reinforcement Learning with Human Feedback (RLHF) ~\citep{Ziegler_Stiennon_Wu_Brown_Radford_Amodei_Christiano_Irving_2019}.  RLHF is applied to update the LLM, guiding it toward outputs that score higher on the reward scale. In the reward model stage, a dataset annotated with human feedback is used to establish the reward, typically scalar in nature, of a particular response. The LLM is then trained to produce responses that will receive higher rewards through a process known as Proximal Policy Optimization (PPO) \citep{Schulman_Wolski_Dhariwal_Radford_Klimov_2017}. This iterative process ensures the model aligns with human expectations, but it can be resource-intensive, requiring significant memory, time, and computational power. 

\begin{figure} [ht]
    \centering
    \includegraphics[width = \textwidth]{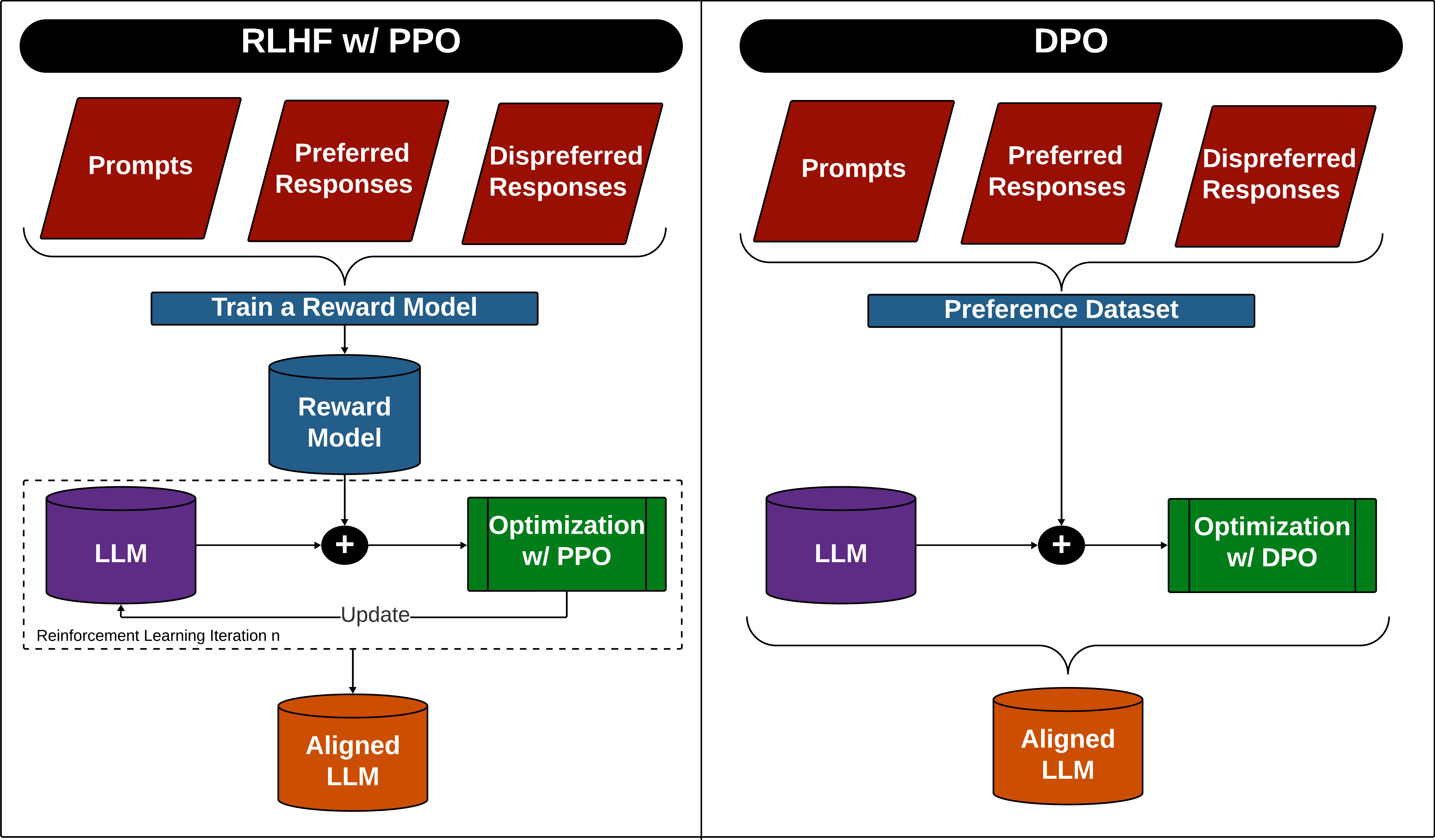}
    \caption{\textbf{Alignment Workflow: PPO v. DPO} An overview of the processes for aligning an LLM through Reinforcement Learning Human Feedback (RLHF) with Proximal Policy Optimization (PPO) and Direct Policy Optimization (DPO).}
    \label{fig:ppo_dpo}
\end{figure}
 
To address these computational challenges, newer paradigms have emerged that streamline Human Alignment training by directly optimizing the LLM-based on human preferences, without the need for a reward model with Direct Preference Optimization (DPO) ~\citep{Rafailov_Sharma_Mitchell_Ermon_Manning_Finn_2023}. DPO reformulates the alignment process into a human-aware loss function (HALO), optimized on a dataset of human preferences where prompts are paired with preferred and dis-preferred responses (Figure ~\ref{fig:ppo_dpo}). This method is particularly promising for aligning LLMs with human preferences and can be applied to ordinal responses, such as the Likert scales commonly seen in human evaluation rubrics. While PPO improves LLM performance by aligning outputs with human preferences, it is often sample-inefficient and can suffer from reward hacking ~\citep{Wen_Zhong_Khan_Perez_Steinhardt_Huang_Boman_He_Feng_2024}. DPO, in contrast, directly optimizes model outputs based on human preferences without needing an explicit reward model, making it more sample-efficient and better aligned with human values. DPO simplifies the training process by focusing directly on the desired outcomes, leading to more stable and interpretable alignment. While these methods have been successfully applied in other domains ~\citep{Cao_Xu_Zhao_Duan_Yang_2024, Iqbal_Mehran_IEEE_2022, Sun_Zhou_Hao_Fan_Lu_Ma_Shen_Guo_2023}, their use in the medical field is under-explored. Training data from the human evaluation rubric on a much smaller scale to overcome labor constraints can be incorporated into a loss function designed for human alignment using DPO. 

\begin{figure} [ht]
    \centering
    \includegraphics[width = \textwidth]{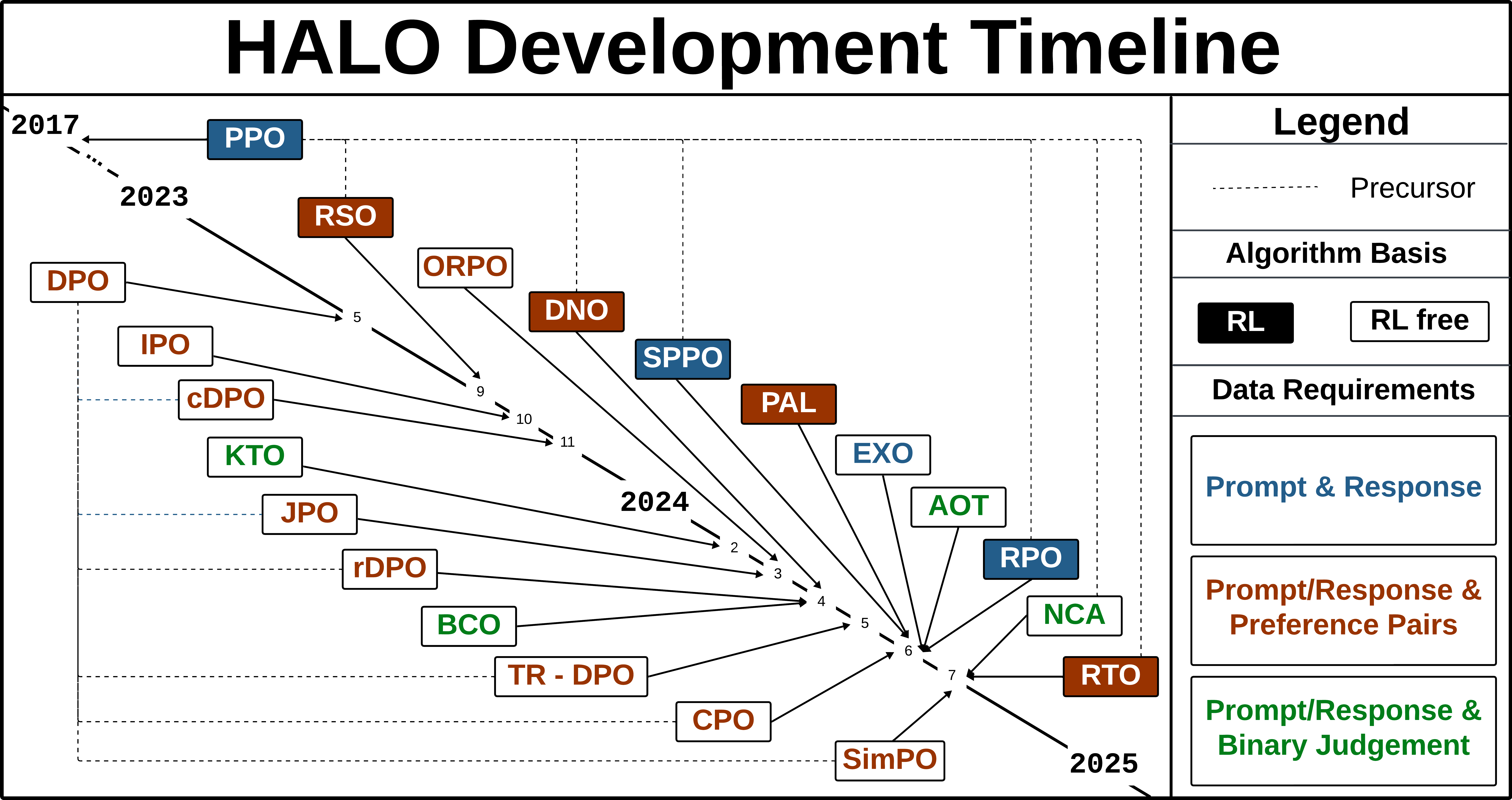}
    \caption{\textbf{Human Aware Loss Functions (HALOs) from PPO to Present} The development timeline for HALOs from the advent of Proximal Policy Optimization (PPO) in 2017 through 2024. HALOs are classified on an algorithmic basis and on their data requirements.\protect\footnotemark[2]
    }
    \label{fig:dpo_var}
\end{figure}

In the last year, many variants of DPO have emerged for alignment training methods that can prevent over-fitting and circumvent DPO's modeling assumptions with modifications to the underlying model and loss function (Figure ~\ref{fig:dpo_var}). Alternative methods such as Joint Preference Optimization (JPO) ~\citep{Bansal_Suvarna_Bhatt_Peng_Chang_Grover_2024} and Simple Preference Optimization (SimPO) ~\citep{Meng_Xia_Chen_2024} were derived from DPO. These methods introduce regularization terms and modifications to the loss function to prevent premature convergence and ensure more robust alignment over a broader range of inputs. Other alternative methods such as Kahneman-Tversky Optimization (KTO) ~\citep{Ethayarajh_Xu_Muennighoff_Jurafsky_Kiela_2024} and Pluralistic Alignment Framework (PAL) ~\citep{Rosset_Cheng_Mitra_Santacroce_Awadallah_Xie_2024} use alternatives to the Bradley-Terry preferences model that underlies DPO. The alternative modeling assumptions used in these methods can prevent the breakdown of DPO's alignment in situations without direct preference data and heterogeneous human preferences.

\footnotetext[2]{The figure includes 
PPO: Proximal Policy Optimization ~\citep{Schulman_Wolski_Dhariwal_Radford_Klimov_2017}, 
DPO: Direct Preference Optimization~\citep{Rafailov_Sharma_Mitchell_Ermon_Manning_Finn_2023}, 
RSO: Statistical Rejection Sampling~\citep{Liu_Zhao_Joshi_Khalman_Saleh_Liu_Liu_2024}, 
IPO: Identity Preference Optimization~\citep{Azar_Rowland_Piot_Guo_Calandriello_Valko_Munos_2023}, 
cDPO: Conservative DPO~\citep{cDPO},
KTO: Kahneman Tversky Optimization~\citep{Ethayarajh_Xu_Muennighoff_Jurafsky_Kiela_2024}, 
JPO: Joint Preference Optimization~\citep{Bansal_Suvarna_Bhatt_Peng_Chang_Grover_2024}, 
ORPO: Odds Ratio Preference Optimization~\citep{Hong_Lee_Thorne_2024}, 
rDPO: Robust DPO~\citep{Chowdhury_Kini_Natarajan_2024}, 
BCO: Binary Classifier Optimization~\citep{Jung_Han_Nam_On_2024}, 
DNO: Direct Nash Optimization~\citep{Rosset_Cheng_Mitra_Santacroce_Awadallah_Xie_2024}, 
TR-DPO: Trust Region DPO~\citep{Gorbatovski_Shaposhnikov_Malakhov_Surnachev_Aksenov_Maksimov_Balagansky_Gavrilov_2024}, 
CPO: Contrastive Preference Optimization~\citep{Xu_Sharaf_Chen_Tan_Shen_Van_Durme_Murray_Kim_2024}, 
SPPO: Self-Play Preference Optimization~\citep{Wu_Sun_Yuan_Ji_Yang_Gu_2024}, 
PAL: Pluralistic Alignment Framework~\citep{Rosset_Cheng_Mitra_Santacroce_Awadallah_Xie_2024}, 
EXO: Efficient Exact Optimization~\citep{Ji_Lu_Niu_Ke_Wang_Zhu_Tang_Huang_2024}, 
AOT: Alignment via Optimal Transport~\citep{Melnyk_Mroueh_Belgodere_Rigotti_Nitsure_Yurochkin_Greenewald_Navratil_Ross_2024}, 
RPO: Iterative Reasoning Preference Optimization~\citep{Pang_Yuan_Cho_He_Sukhbaatar_Weston_2024}, 
NCA: Noise Contrastive Alignment~\citep{Chen_He_Yuan_Cui_Su_Zhu_2024}, 
RTO: Reinforced Token Optimization~\citep{Zhong_Feng_Xiong_Cheng_Zhao_He_Bian_Wang_2024}, 
SimPO: Simple Preference Optimization~\citep{Meng_Xia_Chen_2024}}
\subsection{Drawbacks of LLMs as Evaluators}

LLMs hold promise for automating evaluation, but as with other automated evaluation methods, there are significant challenges to consider. One major issue is the rapid pace at which LLMs and their associated training strategies have evolved. This rapid development often outpaces the ability to thoroughly validate LLM-based evaluators before they are used in practice. In some cases, new optimization techniques are introduced before their predecessors have undergone peer review, and these advancements may lack sufficient mathematical justification. The speed of LLM evolution can make it difficult to allocate time and resources for proper validation, which can compromise their reliability.

Moreover, despite their advancements, LLMs remain sensitive to the prompts and inputs they receive. As LLMs continue to update and change their internal knowledge representations and as their prompts also change, the output can be highly variable. The exact LLM, or model version, that is used can also add another layer of variability. The same prompts and inputs can produce different results based on the LLM's internal structure and pre-training schema. LLMs have also been noted for egocentric bias which could affect evaluations as more and more LLM generated text appears in source texts ~\citep{Koo_Lee_Raheja_Park_Kim_Kang_2024}. As a result, the use of LLMs as evaluators must be accompanied by stringent testing and safety checks to mitigate risks. Ensuring fairness in their responses is also critical, particularly in sensitive domains like healthcare, where biased or stigmatizing language could have serious consequences. These challenges highlight the need for continuous evaluation, testing, and refinement to make LLM-based evaluators both reliable and safe for medical evaluations.

\section{Evaluation Needs for the Clinical Domain}

The development of reliable evaluation strategies is becoming increasingly important as the pace of innovation in GenAI outstrips the speed at which these technologies are validated. In healthcare, the focus on clinical safety must also contend with the time constraints placed on healthcare professionals. While human evaluation rubrics offer a high degree of reliability and accuracy, they are significantly limited by the time commitment required from medical professionals serving as evaluators. Ironically, the technologies being evaluated often aim to reduce the cognitive load on these same professionals, yet they demand further time investment for their performance evaluation.

Automated evaluations, if properly designed for the clinical domain, present a promising alternative to human evaluations. However, traditional non-LLM automated evaluations have thus far fallen short, failing to consistently match the rigor of human evaluation rubrics ~\citep{Croxford_Gao_Patterson_To_Tesch_Dligach_Mayampurath_Churpek_Afshar_2024,Sai_Mohankumar_Khapra_2023}. These metrics frequently overlook hallucinations, fail to assess reasoning quality, and struggle to determine the relevance of generated texts. As LLMs are introduced as potential alternatives for human evaluators, it is critical to consider the unique requirements of the clinical domain. A well-designed LLM evaluator—an "LLM-as-a-judge"—could potentially combine the high reliability of human evaluations with the efficiency of automated methods, while avoiding the pitfalls that have limited existing automated metrics. If executed effectively, such LLM-based evaluations could offer the best of both worlds, ensuring clinical safety without sacrificing the quality of assessments.

\section{Acknowledgements}
We thank Anne Glorioso, Leslie Christensen, and Paije Wilson for their assistance with database selection and search query assistance as UW-Madison subject librarians. EC was supported by an NLM training grant to the Computation and Informatics in Biology and Medicine Training Program (NLM 5T15LM007359). YG was supported by NIH/NLM R00 LM014308-02 and MA was supported by R01LM012973.

\section{Author Contributions}

EC, YG, NP, EF, BP, and MA conceptualized the review. EC completed the literature search and prepared the first draft of the manuscript. EC, YG, NP, EF, KW, BP, and MA contributed to the revision of the manuscript. All authors read through and approved the final manuscript.

\section{Competing Interests}

The authors declare no competing interests.

\begin{center}
    \bibliography{References} %Limit of 60
\end{center}

\end{document}